\documentclass[runningheads]{llncs}
\usepackage{graphicx}
\usepackage{multirow}
\usepackage[dvipsnames]{xcolor}
\usepackage{graphicx}
\usepackage{ifsym}
 \usepackage{mathptmx}      
\usepackage{amsmath}
\usepackage{amssymb}

\usepackage{times}  
\usepackage{helvet} 
\usepackage{courier}  
\usepackage[hyphens]{url}  
\usepackage{graphicx} 

\usepackage{caption} 
\usepackage{colortbl}
\usepackage{nicematrix}
\usepackage{subfig}
\usepackage{amsmath}
\usepackage{amssymb}
\usepackage{booktabs}
\usepackage{xcolor}
\usepackage{tabularx}

\begin{document}
\title{Deep Ear Biometrics for Gender Classification }
%

%
\author{Ritwiz Singh\inst{1}\and
Keshav Kashyap\inst{1}\and
Rajesh Mukherjee\inst{1}\and \\
Asish Bera\inst{2}\and
Mamata Dalui Chakraborty\inst{3}}

\authorrunning{ R. Singh et al.}
\institute{Department of Computer Science and Engineering, Haldia Institute of Technology,  WB \and
Department of Computer Science and Information Systems, Birla Institute of Technology and Science Pilani, Pilani Campus, Rajasthan 333 031, India \and Dept. of Computer Science and Engineering, National Institute of Technology, Durgapur, West Bengal, India \\
\email{ singhritwiz@gmail.com, k.kashyap.aj@gmail.com, rajeshmukherjee.cse@gmail.com, asish.bera@pilani.bits-pilani.ac.in, mamata.dalui@cse.nitdgp.ac.in }}
\maketitle              
\begin{abstract}
Human gender classification based on biometric features is a major concern for computer vision due to its vast variety of applications. The human ear is popular among researchers as a soft biometric trait, because it is less affected by age or changing circumstances, and is non-intrusive. In this study, we have developed a deep convolutional neural network (CNN) model for  automatic gender classification using the samples of ear images. The  performance is evaluated using four cutting-edge pre-trained CNN models. In terms of trainable parameters, the proposed technique requires significantly less computational complexity. The proposed model has achieved 93\% accuracy on the EarVN1.0 ear dataset.

\end{abstract}
\section{Introduction}
Gender identification has become a major concern due to its vast variety of applications, including social communication and connection, commercial visual supervision, banking transactions, illness prognosis, demographic data collection, artificial intelligence (AI) based user interface for customization, consumer analysis for business growth, and many more \cite{kamboj2022comprehensive}, \cite{yaman2019multimodal}. 
Biometric traits have proven their suitability in gender classification as they are non-intrusive, remain invariant with time, and are less influenced by emotions, and circumstances. Several approaches have been proposed using various biometrics traits and their fusion. Biometrics can be broadly categorized into the conventional (aka hard) biometrics (e.g., Face, Hand, Palmprint, etc.) \cite{kar2020lmzmpm}, \cite{bera2021two}; and soft biometrics (e.g., Ear, Gait, Signature, etc.) \cite{yaman2018age},  \cite{mukherjee2021human}.

 In classifying human gender, based on  the iris samples, Bansal and Sharma \cite{bansal2012svm} have employed wavelet transforms and statistical methods. A technique to identify the gender of complete body photographs using part-based gender recognition is proposed in \cite{cao2008gender}. To determine the dialect and gender from parlance data, supervised machine-learning techniques have been employed \cite{hacohen2017language}. Tapia et al. \cite{tapia2017gender} used the local binary patterns (LBP) and Histograms of Gaussian (HOG) descriptors for gender identification from iris images. Gait-based gender categorization methods are suggested by Yu et al. \cite{yu2009study} and Li et al. \cite{li2008gait}. Shan \cite{shan2012learning} has used real-world facial pictures and the LBP description to categorize gender. A novel patch-based LBP with adjustable weights is presented by Chen and Jeng \cite{chen2020new} for the classification of gender on the person's face. 
 
Moreover, a human ear is an unique evidence among the existing biometric traits for
gender identification. The anatomy of the human ear is regarded to be equally
significant as compared to other biometric traits like the face, iris, and gait, for
identifying an individual. Additionally, it is less affected by aging when compared to other biometric traits like the face, \cite{iannarelli1989forensic} and gait \cite{li2008gait}. Other biometric traits may change with age or changing circumstances, but the ear simply scales up.  Numerous methods for ear identification have been proposed. Anwar et al. \cite{anwar2015human}, use geometrical information from human ear images for identification. Various image descriptors have been studied including LPQ, BSIF, LBP, HOG, and surface descriptors in  \cite{pflug2014comparative}. Gender classification from ear images has also been studied in \cite{yaman2018age,gnanasivam2013gender,khorsandi2013gender,emervsivc2017ear}.

However, the effectiveness of ConvNet for gender identification is yet to be  researched thoroughly. In this study, we suggest a basic yet successful CNN-based model that is computationally lightweight. For the experiment, we have considered the EarVN1.0 large-scale ear dataset and achieved significantly good accuracy.

\section{Related Work} \label{relwork}
Existing works on gender identification  can be divided into three types: appearance-based approaches \cite{cao2008gender,basha2012face,shan2012learning,bekios2014robust,hadid2009combining,ueki2004method}, biological feature-based approaches \cite{shue2008role,carrier2001effects,tripathy2012gender,thomas2007learning,bera2014person}, and interpersonal information-based approaches \cite{corney2002gender,liwicki2007automatic,mukherjee2010improving}. External traits of an individual, such as the face, eye brows, hands, fingernails, attire, and footwear, are focused on an appearance-based approach. The biological feature-based techniques make use of biometric characteristics (e.g., fingerprint, ear, iris, etc.) and bio-signals to determine gender (ECG, EEG, etc.). The interpersonal information-based strategy use email, blogs, handwriting, and other types of interpersonal information.

Nowadays, the research community has recently demonstrated a growing interest in biological feature-based techniques approaches since unique biological information of individuals does not change over long periods of time, is non-intrusive, is less influenced by emotions, and is appropriate for applications requiring real-time recognition. \cite{kumar2012automated}. Because of various distinctive features, the ear has gained increased interest in biometric research for identification. 
Several works on gender categorization based on ear biometrics have been conducted.
. In \cite{gnanasivam2013gender}, the authors considered measurements between the ear hole
within the ear hole and seven ear features computed from the masked ear image for identification. Over 342 samples, they examined KNN, Bayes, and neural networks for classification accuracy. KNN performed the best, with a classification accuracy rate of 90.42\%. Histogram intersection
kernel are employed to classify profile face and ear images in \cite{zhang2011hierarchical}. To increase accuracy, they used Bayesian analysis to achieve score-level fusion. The trials were carried out using 2D images. Face and ear attained 95.43\% and 91.78\% perfection, respectively, while fusion achieved 97.68\% precision. Yaman et.  al \cite{yaman2019multimodal}  shows that, much higher accuracy in gender identification ( 99\%) can be achieved through fusion with facial profiles. Authors in \cite{khorsandi2013gender} suggested a sparse representation-based method for identifying gender from ear images. They added Gabor filters for feature extraction to enhance accuracy. The authors in \cite{lei2013gender} have investigated the use of 3D ear images for gender identification. The authors have exploited curvature and indexed shapes of ears for the classification.
In, \cite{yaman2018age},  authors, with the aid of geometrical and appearance-based features identified age, gender, and other biometric traits. A methodology to identify gender and kinship using geometric aspects of ear pictures has been developed in \cite{meng2019gender}. The proposed model has an accuracy of 67.2\% in gender classification.

In this study, a straightforward, but, effective CNN model is proposed and tested for gender classification using the EarVN1.0 \cite{hoang2019earvn1} ear dataset, and performance is evaluated in contrast with four cutting-edge models.

\section{Methodology} \label{method}

\begin{figure}
\centering
\vspace{-0.4 cm}
\includegraphics[scale =0.7]{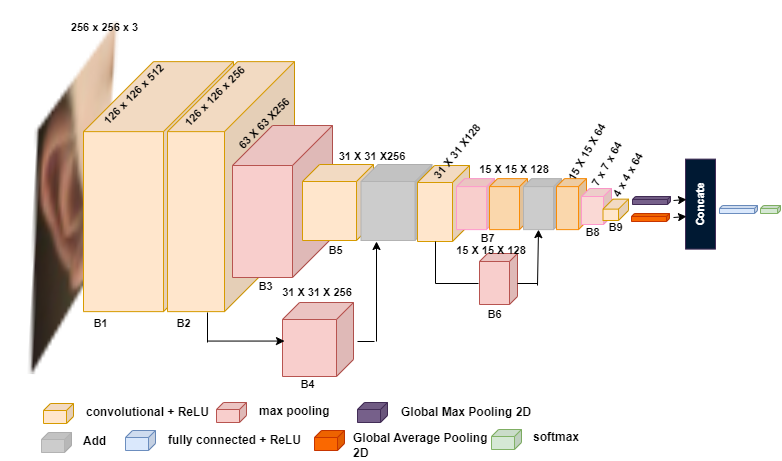}
\caption{Proposed deep model model for Ear biometrics. } \label{model}
\vspace{-0.4 cm}
\end{figure}

A novel CNN-based architecture is proposed for gender detection using ear images to increase classification accuracy and reduce model size. The proffered model's architecture is portrayed in Fig.\ref{model}, and Table \ref{tab0} displays the output shape and parameters for each layer of the proposed model. Let, the input color image  is $\mathtt{X \in I^{h \times w\times 3}}$, $\mathtt{h}$, $\mathtt{w}$, 3 represents height, width and the number of channels. A sequel of convolutional extraction blocks $\mathtt{C(\sum_{n} {\omega_{n}}.{\textbf{x}_n}+b)}$ (where $\omega_n$, $x_n$, and $b$ are weight, input, and bias at $n^{th}$ layer) is employed for the extraction of deep feature-map $ \mathtt{FM=C(X)} \in \mathtt{I}^{h \times w\times c}$ from $\mathtt{X}$. The resultant feature-map is then split into two channels, one for global average pooling $\mathtt{Pool_{avg}}$ (GAP) and another for global max pooling $\mathtt{Pool_{max}}$ (GMP), in order to decrease the spatial aspects by identifying key features along both the pathways, $ FM_a=\mathtt{\mathtt{Pool_{avg}(FM)}}$ and $FM_m=\mathtt{\mathtt{Pool_{max}(FM)}}$. It retains translation invariance while providing a summary of the feature maps. Then a high-level feature map is created by fusing these two feature maps $\mathtt{FM_c=FM_a \oplus FM_m}$, $\oplus$ represents linear concatenation function. And lastly, the \textit{softmax} estimates the likelihood ($\rho$) of gender classification from the input image, $\mathtt{\rho [0, 1]=softmax(FM_c)}$. '0' is designated as the 'female' class in the binary classification issue, while '1' is designated as the 'male' class.
\begin{equation} 
\centering
 \mathtt{FM=C(X)};  \ \ \textrm{ where} \ \  \mathtt{FM} \in \mathtt{I}^{h \times w\times c};  \ \ \mathtt{X} \in \mathtt{I}^{h \times w\times 3}; \ \ \textrm{ and} \ \ \mathtt{C(\sum_{n} {\omega_{n}}.{\textbf{x}_n}+b)}  \ \ \textrm{ is a CNN}
\end{equation}
\begin{equation} 
\centering
FM_a={Pool_{avg}}{(\mathtt{FM})} \ \ \textrm{and} 
\ \ FM_m={Pool_{max}}{(\mathtt{FM})}; \ \  
\end{equation}
\begin{equation}
\centering
 FM_c=[FM_a \oplus FM_m ]; \ \ \textrm{and} \ \  
\rho=softmax(FM_c)
\end{equation}

\begin{table}[]
\centering
\caption{Layer outputs and parameters of proposed model}\label{tab0}
\begin{tabular}{|c|c|c|c|}
\hline
\rowcolor[HTML]{9B9B9B} 
Blks                 & Layer (type)          & Output Shape          & Parameter   \\ \hline
                     & Input Layer          & (None,256,256,3)      & 0       \\ \hline
B1                   & Conv\_2D\_1          & (None, 126, 126, 512) & 38912   \\ \hline
B2                   & Conv\_2D\_2          & (None, 126, 126, 256) & 1179904 \\ \hline
B3                   & Maxpooling\_2D\_1    & (None, 63, 63, 256)   & 0       \\ \hline
B4                   & Maxpooling\_2D\_2    & (None, 31, 31, 256)   & 0       \\ \hline
                     & Conv\_2D\_3          & (None, 31, 31, 256)   & 590080  \\ \cline{2-4} 
                     & Add\_1               & (None, 31, 31, 256)   & 0       \\ \cline{2-4} 
\multirow{-3}{*}{B5} & Conv\_2D\_4          & (None, 31, 31, 128)   & 295040  \\ \hline
B6                   & Maxpooling\_2D\_3    & (None, 15, 15, 128)   & 0       \\ \hline
                     & Maxpooling\_2D\_4    & (None, 15, 15, 128)   & 0       \\ \cline{2-4} 
                     & Conv\_2D\_5          & (None, 15, 15, 128)   & 65664   \\ \cline{2-4} 
                     & Add\_2               & (None, 15, 15, 128)   & 0       \\ \cline{2-4} 
\multirow{-4}{*}{B7} & Conv\_2D\_6          & (None, 15, 15, 64)    & 73792   \\ \hline
B8                   & Maxpooling\_2D\_5    & (None, 7, 7, 64)      & 0       \\ \hline
B9                   & Conv\_2D\_6          & (None, 4, 4, 64)      & 36928   \\ \hline
GAP                  & GlobalAvgPooling\_2D & (None, 64)            & 0       \\ \hline
GMP                  & GlobalMaxPooling\_2D & (None, 64)            & 0       \\ \hline
Concat               & Concatenate          & (None, 128)            & 0       \\ \hline
Dense                & Dense                & (None, 2)             & 258     \\ \hline
\end{tabular}
\end{table}
\noindent \textbf{Implementation:} 
The proposed model comprises a sequel of convolution extraction blocks to extract the feature maps. Next, GAP and GMP are applied independently on the extracted feature map (Fig.\ref{model}). Finally, \textit{softmax} is applied to calculate output probabilities by concatenating the two acquired feature maps. To optimize the training job, binary cross-entropy ($\gamma$) is employed as a loss function. 
\begin{equation} 
\gamma =\frac{1}{n}{\sum_{t=1} ^ {k} - [{y}_{t}.log(P_t)} +(1-{y}_{t}).log(1-P_t) ]\\
\end{equation}
where, ${P_t}$ and $(1-{P_t})$ denote the likelihood of female and male classes, respectively. ${y_t}$ and $\mathtt{\bar{y}_t}$ indicate the actual and predicted class labels, and n represents the number of data samples. With a learning rate of 0.001, the Adaptive Moment Estimation (Adam) optimizer is employed. The model is trained for 100 epochs with 32 mini batch-size. The input shape is $256\times256\times3$, also the suggested model is trained over 100 epochs in 32-batch iterations. The data augmentation process is used to increase the data's size and diversity by including $\pm 0.2$ horizontal flips and $\pm 0.2$ rotations. The output feature dimension ($\mathtt{FM}$)  is $4\times4\times64$. Pooling reduces the size of a feature map
from $w\times h\times c$ to $1\times 1\times c$. Then the feature map is further reduced to 64 ($\mathtt{FM_a=FM_m=64}$) by using the GAP and GMP layers in two distinct channels. Finally, feature maps from both channels (GAP and GMP) are concatenated ($\mathtt{FM_c=128}$). To avoid the overfitting problem, a dropout rate of 0.2 is used in the experiment as regularization. 

\section{Experiments and Result} \label{exp}

%

In this section, We present the outcomes of the proposed model. In order to compare the performance of the proposed model, we have used four base models in which four CNNs namely, VGG19, Xception, ResNet50, and MobileNet have been used as the backbone and tuned with the target dataset. For faster learning convergence, pre-trained ImageNet weights are employed in the considered base models. Features are extracted from the last layers of the backbone CNNs, and then subsequent dense layer is applied for the classification. To assess the model's performance, we have used the EarVN1.0 \cite{hoang2019earvn1} dataset. The dataset contains the large-scale ear images of 164 subjects and a total of 28,412 images.

For assessment, the Receiver Operator Characteristic (ROC) curve is used. AUC, which serves as a summary of the ROC curve, calculates the capacity to differentiate between the classes. In order to discriminate between the male and female categories, a higher AUC is indicative of greater performance. The accuracy (\%) and AUC (\%) metrics are employed.  Model parameters of the CNN models are expressed in millions (M). Furthermore, both accuracy and recall are taken into account in order to properly assess the efficacy of the proposed framework. Higher recall demonstrates a higher likelihood that a specific category may be identified, whereas higher precision suggests the efficiency of a model used to predict a certain category.

\begin{figure}
\subfloat{\includegraphics[width =0.24\textwidth]{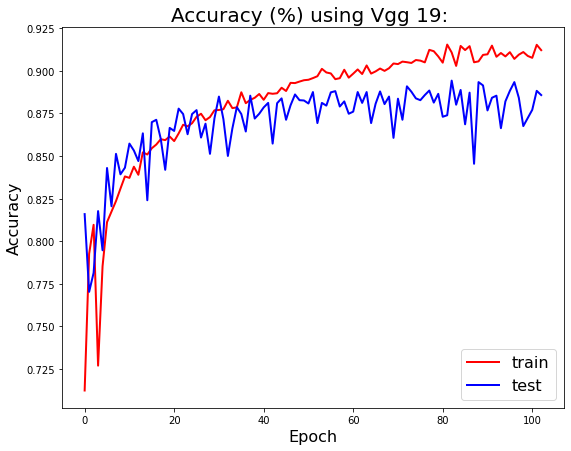}}
\subfloat{\includegraphics[width =0.24\textwidth]{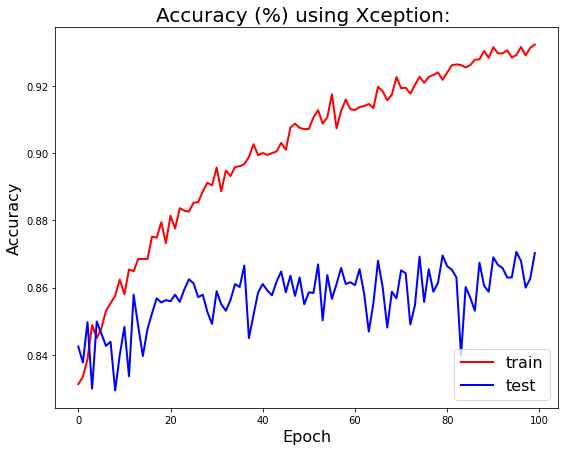}}
\subfloat{\includegraphics[width =0.24\textwidth]{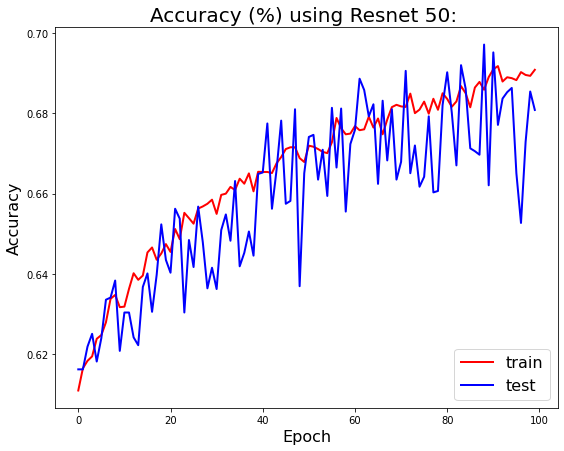}}
\subfloat{\includegraphics[width =0.24\textwidth]{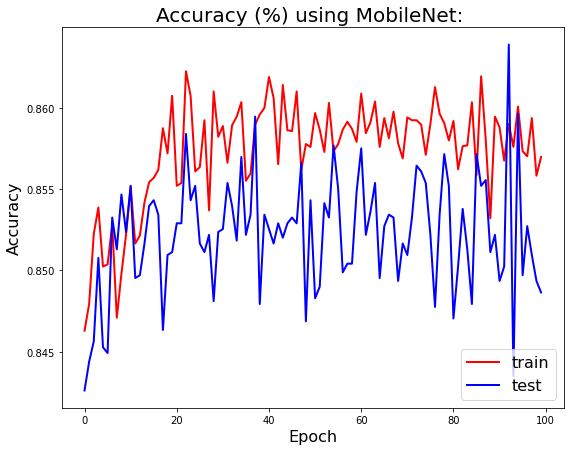}}
\vspace{ -0.3 cm}
\subfloat{\includegraphics[width = 0.24\textwidth]{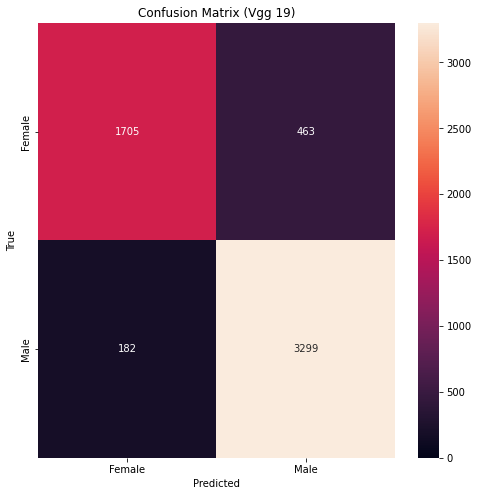}}
\subfloat{\includegraphics[width =0.24\textwidth]{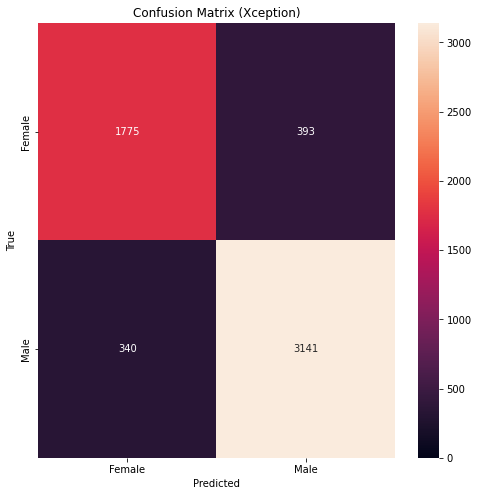}}
\subfloat{\includegraphics[width =0.24\textwidth]{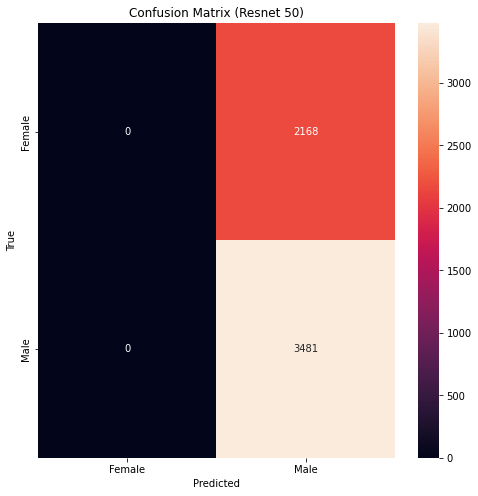}}
\subfloat{\includegraphics[width =0.24\textwidth]{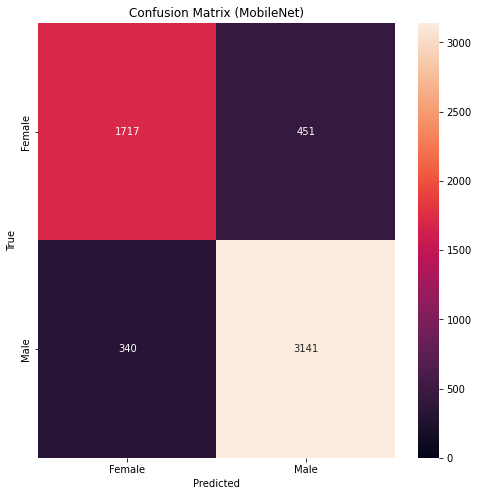}} 
\caption{Accuracy and confusion matrix of VGG-19, Xception,  ResNet-50, and MobileNet (left to right) }
\label{val}
\end{figure}

\begin{figure}
\centering
\subfloat{\includegraphics[width =0.45\textwidth ]{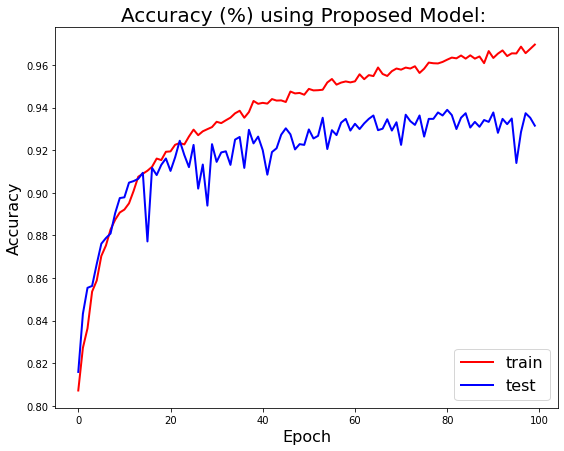}} 
\subfloat{\includegraphics[width =0.35\textwidth]{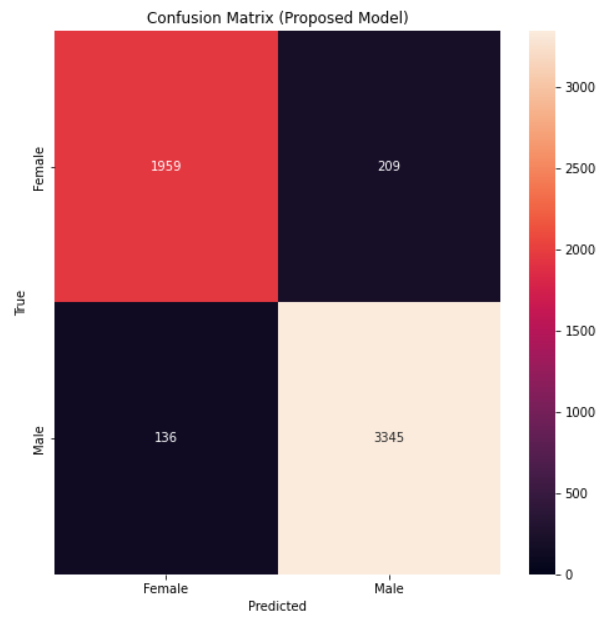}}
\caption{Gender recognition performance of the proposed model }
\label{val1}
\end{figure}

The experiments are conducted on the Kaggle notebook platform equipped with an NVIDIA Tesla T4 GPU. We have tested and compared the performance of all four state-of-the-art pre-trained models and the proposed model on the EarVN1.0\cite{hoang2019earvn1} ear dataset with a 70:30 train-test ratio, and the results are displayed in Table \ref{acc1}. The accuracy indicates that our method  outperforms four well-known state-of-the-art classification models, as stated above. Additionally, it should be highlighted that the computational complexity of the proposed model is considerably lower than the state-of-the-art models. We achieved 93\% accuracy in the proposed model on the EarVN1.0 dataset, whereas only VGG-19 and MobileNet ensure the maximum 85\% accuracy among the examined state-of-the-art models.
The confusion matrix is used to assess a model's utility and envision its performance. In a confusion matrix, diagonal values correspond to predicted labels that are identical to actual labels, whereas non-diagonal values correspond to classifier-erroneously classified labels.
Fig. \ref{val} depicts the confusion matrix and training-testing accuracy during the training process for state-of-the-art models. Fig. \ref{val1} depicts the same for the proposed model.
The test results imply that our proposed deep architecture performs better than state-of-the-art computationally heavy models in the context of precision and recall metrics.     
\vspace{ -0.60 cm}
\begin{table}
\centering
\caption{Efficacy of the considered base models along with the proposed model }\label{tab1}
\begin{tabular}{|l||c|c|c|c|c|c|c|c|}
\hline
{CNN} & \multicolumn{2}{c|}{Female} & \multicolumn{2}{c|}{Male} & {Accuracy} & Model \\
  Model& Precision & recall & Precision & recall &  & Parameters (M)\\
\hline
VGG-19 &  86 & 73   & 84 & 93 &  85  & 20.07 \\
ResNet50 &    67 & 31  & 68 &89 &  68 & 23.59\\
Xception &  78 & 83   & 89 &85 &  84  & 20.86 \\ \hline
MobileNet &  83 & 77 &  86 & 91 & 85  & 3.23 \\
Proposed Model &  \textbf{92} & \textbf{89}   & \textbf{93} & \textbf{95} &  \textbf{93}  & \textbf{2.46} \\
\hline
\end{tabular} \label{acc1}
\vspace{ -0.60 cm}
\end{table}
\vspace{ -0.90 cm}

%

\begin{figure}
\centering
\subfloat{\includegraphics[width =0.24\textwidth]{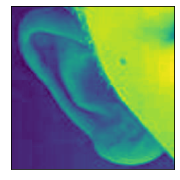}}
\subfloat{\includegraphics[width =0.24\textwidth]{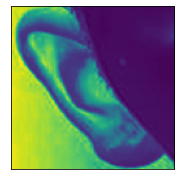}}
\subfloat{\includegraphics[width = 0.24\textwidth]{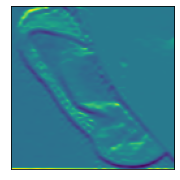}}
\subfloat{\includegraphics[width = 0.24\textwidth]{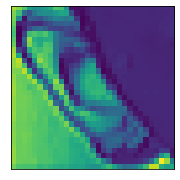}}

\subfloat{\includegraphics[width =0.24\textwidth]{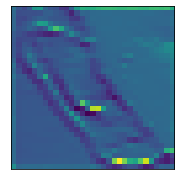}}
\subfloat{\includegraphics[width =0.24\textwidth]{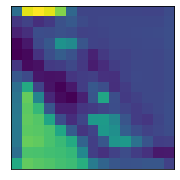}} 
\subfloat{\includegraphics[width =0.24\textwidth]{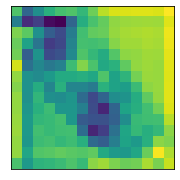}}
\subfloat{\includegraphics[width = 0.24\textwidth]{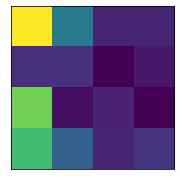}}
\vspace{-0.2 cm}
\caption{ Feature map representation of  proposed model at different convolutional layers. } 
\label{FMap}
\end{figure}

\vspace{-0.2 cm}
\noindent \textbf{Feature Map Representation:}
Fig.\ref{FMap} displays the feature maps from various layers of the proposed model. The low-level features depict an ear shape with earlier levels having more distinct details. On the other hand, the network pipeline's deeper levels depict high-level characteristics with fewer details.

\clearpage
\section{Conclusion}  \label{con}
In this paper, a simple and novel deep CNN method is devised for gender classification using ear biometric samples. The proposed model is tested on EarVN1.0 ear dataset. The proposed model outperforms four computationally demanding state-of-the-art model CNNs, namely VGG-19, Xception, ResNet-50, and MobileNet. Thus, the proposed approach makes the gender categorization more practical. We intend to develop a new deep model for soft biometric features, and a fusion of them, to broaden the method's usefulness for real-world applications.  

\bibliographystyle{splncs04}
\bibliography{Ref_ReviewPaper.bib}

\begin{thebibliography}{10}
\providecommand{\url}[1]{\texttt{#1}}
\providecommand{\urlprefix}{URL }
\providecommand{\doi}[1]{https://doi.org/#1}

\bibitem{anwar2015human}
Anwar, A.S., Ghany, K.K.A., Elmahdy, H.: Human ear recognition using
  geometrical features extraction. Procedia Computer Science  \textbf{65},
  529--537 (2015)

\bibitem{bansal2012svm}
Bansal, A., Agarwal, R., Sharma, R.: Svm based gender classification using iris
  images. In: 2012 fourth international conference on computational
  intelligence and communication networks. pp. 425--429. IEEE (2012)

\bibitem{basha2012face}
Basha, A.F., Jahangeer, G.S.B.: Face gender image classification using various
  wavelet transform and support vector machine with various kernels.
  International Journal of Computer Science Issues (IJCSI)  \textbf{9}(6), ~150
  (2012)

\bibitem{bekios2014robust}
Bekios-Calfa, J., Buenaposada, J.M., Baumela, L.: Robust gender recognition by
  exploiting facial attributes dependencies. Pattern recognition letters
  \textbf{36},  228--234 (2014)

\bibitem{bera2014person}
Bera, A., Bhattacharjee, D., Nasipuri, M.: Person recognition using alternative
  hand geometry. International Journal of Biometrics  \textbf{6}(3),  231--247
  (2014)

\bibitem{bera2021two}
Bera, A., Bhattacharjee, D., Shum, H.P.: Two-stage human verification using
  handcaptcha and anti-spoofed finger biometrics with feature selection. Expert
  Systems with Applications  \textbf{171},  114583 (2021)

\bibitem{cao2008gender}
Cao, L., Dikmen, M., Fu, Y., Huang, T.S.: Gender recognition from body. In:
  Proceedings of the 16th ACM international conference on Multimedia. pp.
  725--728 (2008)

\bibitem{carrier2001effects}
Carrier, J., Land, S., Buysse, D.J., Kupfer, D.J., Monk, T.H.: The effects of
  age and gender on sleep eeg power spectral density in the middle years of
  life (ages 20--60 years old). Psychophysiology  \textbf{38}(2),  232--242
  (2001)

\bibitem{chen2020new}
Chen, W.S., Jeng, R.H.: A new patch-based lbp with adaptive weights for gender
  classification of human face. Journal of the Chinese Institute of Engineers
  \textbf{43}(5),  451--457 (2020)

\bibitem{corney2002gender}
Corney, M., De~Vel, O., Anderson, A., Mohay, G.: Gender-preferential text
  mining of e-mail discourse. In: 18th Annual Computer Security Applications
  Conference, 2002. Proceedings. pp. 282--289. IEEE (2002)

\bibitem{emervsivc2017ear}
Emer{\v{s}}i{\v{c}}, {\v{Z}}., {\v{S}}truc, V., Peer, P.: Ear recognition: More
  than a survey. Neurocomputing  \textbf{255},  26--39 (2017)

\bibitem{gnanasivam2013gender}
Gnanasivam, P., Muttan, S.: Gender classification using ear biometrics. In:
  Proceedings of the Fourth International Conference on Signal and Image
  Processing 2012 (ICSIP 2012). pp. 137--148. Springer (2013)

\bibitem{hacohen2017language}
HaCohen-Kerner, Y., Hagege, R.: Language and gender classification of speech
  files using supervised machine learning methods. Cybernetics and Systems
  \textbf{48}(6-7),  510--535 (2017)

\bibitem{hadid2009combining}
Hadid, A., Pietik{\"a}inen, M.: Combining appearance and motion for face and
  gender recognition from videos. Pattern Recognition  \textbf{42}(11),
  2818--2827 (2009)

\bibitem{hoang2019earvn1}
Hoang, V.T.: Earvn1. 0: A new large-scale ear images dataset in the wild. Data
  in brief  \textbf{27} (2019)

\bibitem{iannarelli1989forensic}
Iannarelli, A.: Forensic identification series: ear identification. Paramont,
  California  \textbf{5} (1989)

\bibitem{kamboj2022comprehensive}
Kamboj, A., Rani, R., Nigam, A.: A comprehensive survey and deep learning-based
  approach for human recognition using ear biometric. The Visual Computer
  \textbf{38}(7),  2383--2416 (2022)

\bibitem{kar2020lmzmpm}
Kar, A., Pramanik, S., Chakraborty, A., Bhattacharjee, D., Ho, E.S., Shum,
  H.P.: Lmzmpm: local modified zernike moment per-unit mass for robust human
  face recognition. IEEE Transactions on Information Forensics and Security
  \textbf{16},  495--509 (2020)

\bibitem{khorsandi2013gender}
Khorsandi, R., Abdel-Mottaleb, M.: Gender classification using 2-d ear images
  and sparse representation. In: 2013 IEEE Workshop on applications of computer
  vision (WACV). pp. 461--466. IEEE (2013)

\bibitem{kumar2012automated}
Kumar, A., Wu, C.: Automated human identification using ear imaging. Pattern
  Recognition  \textbf{45}(3),  956--968 (2012)

\bibitem{lei2013gender}
Lei, J., Zhou, J., Abdel-Mottaleb, M.: Gender classification using
  automatically detected and aligned 3d ear range data. In: 2013 International
  Conference on Biometrics (ICB). pp.~1--7. IEEE (2013)

\bibitem{li2008gait}
Li, X., Maybank, S.J., Yan, S., Tao, D., Xu, D.: Gait components and their
  application to gender recognition. IEEE Transactions on Systems, Man, and
  Cybernetics, Part C (Applications and Reviews)  \textbf{38}(2),  145--155
  (2008)

\bibitem{liwicki2007automatic}
Liwicki, M., Schlapbach, A., Loretan, P., Bunke, H.: Automatic detection of
  gender and handedness from on-line handwriting. In: Proc. 13th Conf. of the
  Graphonomics Society. pp. 179--183. Citeseer (2007)

\bibitem{meng2019gender}
Meng, D., Nixon, M.S., Mahmoodi, S.: Gender and kinship by model-based ear
  biometrics. In: 2019 International Conference of the Biometrics Special
  Interest Group (BIOSIG). pp.~1--5. IEEE (2019)

\bibitem{mukherjee2010improving}
Mukherjee, A., Liu, B.: Improving gender classification of blog authors. In:
  Proceedings of the 2010 conference on Empirical Methods in natural Language
  Processing. pp. 207--217 (2010)

\bibitem{mukherjee2021human}
Mukherjee, R., Bera, A., Bhattacharjee, D., Nasipuri, M.: Human gender
  classification based on hand images using deep learning. Tech. rep.,
  EasyChair (2021)

\bibitem{pflug2014comparative}
Pflug, A., Paul, P.N., Busch, C.: A comparative study on texture and surface
  descriptors for ear biometrics. In: 2014 International carnahan conference on
  security technology (ICCST). pp.~1--6. IEEE (2014)

\bibitem{shan2012learning}
Shan, C.: Learning local binary patterns for gender classification on
  real-world face images. Pattern recognition letters  \textbf{33}(4),
  431--437 (2012)

\bibitem{shue2008role}
Shue, Y.L., Iseli, M.: The role of voice source measures on automatic gender
  classification. In: 2008 IEEE International Conference on Acoustics, Speech
  and Signal Processing. pp. 4493--4496. IEEE (2008)

\bibitem{tapia2017gender}
Tapia, J., Viedma, I.: Gender classification from multispectral periocular
  images. In: 2017 IEEE international joint conference on biometrics (IJCB).
  pp. 805--812. IEEE (2017)

\bibitem{thomas2007learning}
Thomas, V., Chawla, N.V., Bowyer, K.W., Flynn, P.J.: Learning to predict gender
  from iris images. In: 2007 First IEEE International Conference on Biometrics:
  Theory, Applications, and Systems. pp.~1--5. IEEE (2007)

\bibitem{tripathy2012gender}
Tripathy, R.K., Acharya, A., Choudhary, S.K.: Gender classification from ecg
  signal analysis using least square support vector machine. American Journal
  of Signal Processing  \textbf{2}(5),  145--149 (2012)

\bibitem{ueki2004method}
Ueki, K., Komatsu, H., Imaizumi, S., Kaneko, K., Sekine, N., Katto, J.,
  Kobayashi, T.: A method of gender classification by integrating facial,
  hairstyle, and clothing images. In: Proceedings of the 17th International
  Conference on Pattern Recognition, 2004. ICPR 2004. vol.~4, pp. 446--449.
  IEEE (2004)

\bibitem{yaman2018age}
Yaman, D., Eyiokur, F.I., Sezgin, N., Ekenel, H.K.: Age and gender
  classification from ear images. In: 2018 International Workshop on Biometrics
  and Forensics (IWBF). pp.~1--7. IEEE (2018)

\bibitem{yaman2019multimodal}
Yaman, D., Irem~Eyiokur, F., Kemal~Ekenel, H.: Multimodal age and gender
  classification using ear and profile face images. In: Proceedings of the
  IEEE/CVF Conference on Computer Vision and Pattern Recognition Workshops.
  pp.~0--0 (2019)

\bibitem{yu2009study}
Yu, S., Tan, T., Huang, K., Jia, K., Wu, X.: A study on gait-based gender
  classification. IEEE Transactions on image processing  \textbf{18}(8),
  1905--1910 (2009)

\bibitem{zhang2011hierarchical}
Zhang, G., Wang, Y.: Hierarchical and discriminative bag of features for face
  profile and ear based gender classification. In: 2011 International joint
  conference on biometrics (IJCB). pp.~1--8. IEEE (2011)

\end{thebibliography}
\end{document}